\documentclass[letterpaper]{article} 
\usepackage{aaai25}  
\usepackage{times}  
\usepackage{helvet}  
\usepackage{courier}  
\usepackage[hyphens]{url}  
\usepackage{graphicx} 
\usepackage{natbib}  
\usepackage{caption} 
\usepackage{algorithm}
\usepackage{algorithmic}
\usepackage{xcolor}

\usepackage{newfloat}
\usepackage{listings}
\DeclareCaptionStyle{ruled}{labelfont=normalfont,labelsep=colon,strut=off} 
\floatstyle{ruled}
\newfloat{listing}{tb}{lst}{}
\floatname{listing}{Listing}
\usepackage{algorithm}
\usepackage{algorithmic}

\usepackage{pifont}

\usepackage{algorithm}
\usepackage{algorithmic}
\usepackage{float}
\usepackage{multirow}
\usepackage{booktabs}
\usepackage{newtxtext}
\usepackage{bm}
\usepackage{float}
\usepackage{subfigure}
\usepackage{multicol}
\usepackage{pdfpages}
\usepackage{indentfirst}
\usepackage{diagbox}
\usepackage{amsmath}
\usepackage{amsthm}

\usepackage{newfloat}
\usepackage{listings}
\floatstyle{ruled}
\newfloat{listing}{tb}{lst}{}
\floatname{listing}{Listing}
\usepackage{cite}

\usepackage{url}

\usepackage{amssymb}
\usepackage{amsmath}
\usepackage{mathdots} 
\usepackage{microtype} 
\usepackage{array}
\usepackage{threeparttable}
\usepackage{array}
\usepackage{color}
\makeatletter
\newcommand{\thickhline}{%
    \noalign {\ifnum 0=`}\fi \hrule height 2pt
    \futurelet \reserved@a \@xhline
}
\usepackage{hyperref}

\usepackage{adjustbox}
\usepackage{amssymb}
\usepackage{natbib}

\usepackage{newfloat}
\usepackage{listings}
\DeclareCaptionStyle{ruled}{labelfont=normalfont,labelsep=colon,strut=off} 
\floatstyle{ruled}
\newfloat{listing}{tb}{lst}{}
\floatname{listing}{Listing}
\newcommand{\cmark}{\ding{51}}%
\title{Efficient 3D Recognition with Event-driven Spike Sparse Convolution}
\author{
     Xuerui Qiu\textsuperscript{\rm 1,2},
    Man Yao\textsuperscript{\rm 1}\thanks{Corresponding author.},
    Jieyuan Zhang\textsuperscript{\rm 3},
    Yuhong Chou\textsuperscript{\rm 1,4},
    Ning Qiao\textsuperscript{\rm 5}
    Shibo Zhou\textsuperscript{\rm 6},
    Bo Xu\textsuperscript{\rm 1},
    Guoqi Li$^{1*}$
}
\affiliations{
    \textsuperscript{\rm 1}Institute of Automation, Chinese Academy of Sciences\\
    \textsuperscript{\rm 2}School of Future Technology, University of Chinese Academy of Sciences\\
    \textsuperscript{\rm 3}University of Electronic Science and Technology of China,
    \textsuperscript{\rm 4}The Hong Kong Polytechnic University\\

\textsuperscript{\rm 5}SynSense AG Corporation,\textsuperscript{\rm 6}Huinao Zhixin \\
}

\usepackage{bibentry}

\begin{document}

\maketitle
\begin{abstract}
Spiking Neural Networks (SNNs) provide an energy-efficient way to extract 3D spatio-temporal features. Point clouds are sparse 3D spatial data, which suggests that SNNs should be well-suited for processing them. However, when applying SNNs to point clouds, they often exhibit limited performance and fewer application scenarios. We attribute this to inappropriate preprocessing and feature extraction methods. To address this issue, we first introduce the Spike Voxel Coding (SVC) scheme, which encodes the 3D point clouds into a sparse spike train space, reducing the storage requirements
and saving time on point cloud preprocessing. Then, we propose a Spike Sparse Convolution (SSC) model for efficiently extracting 3D sparse point cloud features. Combining SVC and SSC, we design an efficient 3D SNN backbone  (E-3DSNN), which is friendly with neuromorphic hardware. For instance, SSC can be implemented on neuromorphic chips with only minor modifications to the addressing function of vanilla spike convolution. Experiments on ModelNet40, KITTI, and Semantic KITTI datasets demonstrate that E-3DSNN achieves state-of-the-art (SOTA) results with remarkable efficiency. 
Notably, our E-3DSNN (1.87M) obtained 91.7\% top-1 accuracy on ModelNet40, surpassing the current best SNN baselines (14.3M) by 3.0\%.  To our best knowledge, it is the first direct training 3D SNN backbone that can simultaneously handle various 3D computer vision tasks (e.g., classification, detection, and segmentation) with an event-driven nature. Code is available \href{https://github.com/bollossom/E-3DSNN/}{here}.
\end{abstract}
\section{Introduction}
3D recognition has been a highly researched area due to its wide applications in autonomous driving \cite{cui2021deep}, virtual reality \cite{zhu2024advancements}, and robotics \cite{pomerleau2015review}. 
However, these methods involve numerous operations, leading to high computational costs and energy consumption, which limits their deployment on resource-constrained devices.
Bio-inspired Spiking Neural Networks (SNNs) provide an energy-efficient way to extract features from 3D event streams due to their unique event-driven nature and spatio-temporal dynamics \cite{maass1997networks,roy2019towards,li2023brain}. For instance, the Speck \cite{yao2024spike} chip uses event-by-event sparse processing to handle event streams, with operational power consumption as low as 0.7 mW. Point clouds and event streams are both sparse 3D data, which theoretically suggests that SNNs should be well-suited for processing 3D sparse point clouds.  However, when applying SNNs to point clouds, they often exhibit limited performance and fewer application scenarios in most cases.
 \par
 For instance, most applications of SNN algorithms \cite{lan2023efficient,wu2024spikepoint,ren2024spiking} in 3D recognition are limited to simple 3D classification tasks with toy model datasets \cite{wu20153d} and the performance gap between these works and ANNs is significant. To better apply the SNNs in the efficient 3D recognition field, we identify the key issues in the SNN processing of point clouds. The first is the appropriate preprocessing method.  Vanilla methods \cite{ren2024spiking,wu2024point} use point-based methods to process input point clouds, the inherent sparsity of SNNs can obscure local geometric information, and the high computational load makes training on large datasets time-consuming.
 The second is selecting efficient feature extraction tools. 3D data itself has high redundancy. While SNNs use 2D spike convolution effectively for event streams, applying it to 3D sparse point clouds results in cubic growth in computational complexity as it calculates each point. This makes feature extraction inefficient and challenging.
 \par
 \par
 \begin{figure*}[!t]
\centering
\includegraphics[scale=0.312]{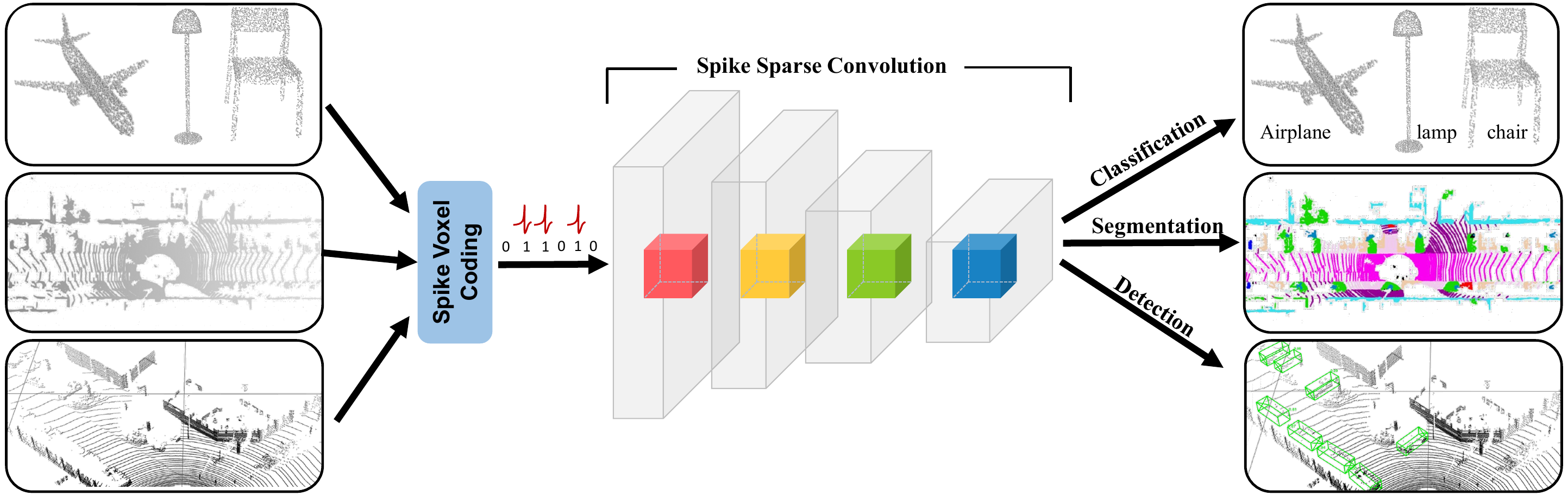}
\vspace{-1mm}
\caption{The workflow of our efficient 3D SNN backbone (E-3DSNN), which uses residual connections between membrane potentials and handles various 3D computer vision tasks with only sparse ACcumulate. It consists of two main components:  the Spike Voxel Coding (SSC) and  Spike Sparse Convolution (SSC). The SVC scheme first voxelizes the input 3D points. Then, the voxelized data is transformed into spatio-temporal spike trains using sparse convolution and spiking neurons. The SSC block only calculates the overlapping activation features between the center of the point cloud and the convolution kernel.}
\label{fig:main}
\end{figure*}

\begin{figure}[t]
    \centering
    \includegraphics[width=1\linewidth]{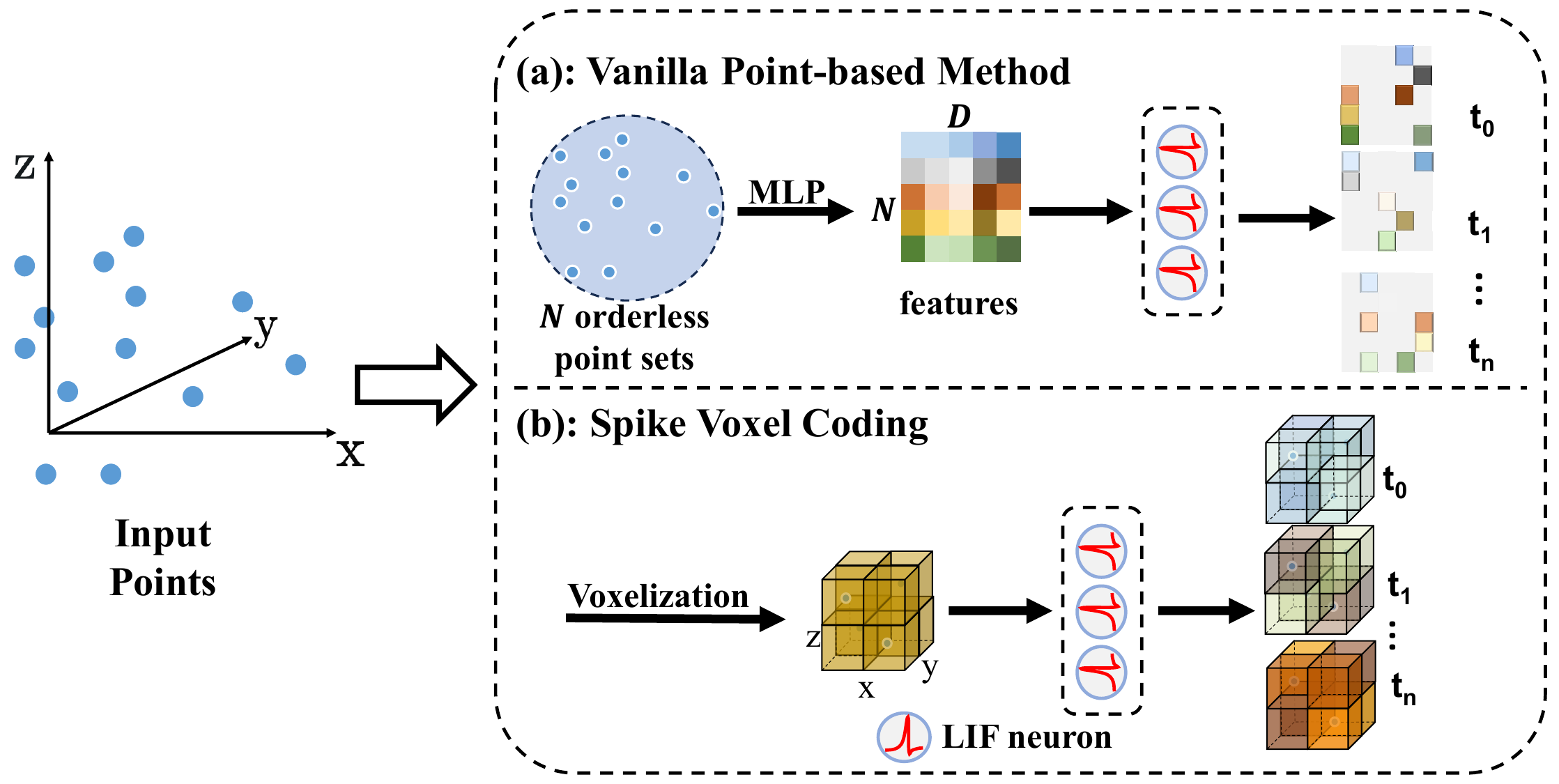}
    \caption{Comparison of different point cloud pre-processing ways in SNN. (a) The vanilla point-based method \cite{lan2023efficient,ren2024spiking,wu2024spikepoint} directly processes raw points, but the inherent sparsity of SNNs can obscure local geometric details.
(b) We proposed a spike voxel coding (SVC) scheme, which leverages the sparsity of SNNs and, after additional voxelization pretreatment, can handle structural data with higher efficiency and lower power consumption.}
    \label{fig:svc}
\end{figure}

  In this work, we aim to address the above issues in a unified manner.
 Our goal is to highlight the low power consumption and distinct sparse event-driven advantages of SNNs. We accomplish this through two main approaches.
 First, we propose a Spike Voxel Coding (SVC) scheme for processing point clouds. As shown in Fig. \ref{fig:svc}, SVC can encode the 3D point clouds into sparse and spatio-temporal spike trains, reducing the storage requirements and saving time on point cloud preprocessing.
 Second, we propose a Spike Sparse Convolution (SSC) block for extracting 3D point cloud features, which leverages sparsity to reduce redundant computations on background points and avoids the densification issues associated with vanilla spike convolution. As shown in Fig. \ref{fig:sparseconv}, SSC adds just one extra condition compared to the Vanilla Spike Convolution (VSC). Because of their similarities, SSC can be implemented on neuromorphic chips with only minor modifications to the VSC addressing function. 
 Finally, leveraging SVC and SSC, we redesigned an efficient 3D SNN backbone (E-3DSNN) using residual connections between membrane potentials, as shown in Fig. \ref{fig:main}.
 To demonstrate the effectiveness of E-3DSNN, we evaluate our models on the simple  ModelNet40 \cite{wu20153d} and two large-scale benchmarks (e.g., KITTI \cite{KITTI} and Semantic KITTI \cite{behley2019semantickitti} datasets). E-3DSNN achieves leading performance with high efficiency with only sparse ACcumulate (AC) in SNNs.
Our main contribution can be summarized as:
\begin{itemize}
\item We introduce the SVC scheme and SSC block, enhancing SNN efficiency and performance in processing 3D point clouds. SVC converts point clouds to sparse spike trains, while SSC extracts effective representations from them.
\item We explore suitable architectures in SNNs and redesigned LiDAR-based 3D SNN backbone by residual connections between membrane potentials, handling various 3D computer vision tasks with sparse AC operation.
\item Experiments demonstrate that our E-3DSNN achieves outstanding accuracy with remarkable efficiency up to 11$\times$ on various datasets (e.g., ModelNet40, KITTI, and semantic KITTI), revealing the potential of SNNs in efficient 3D recognition. 
\end{itemize}

 



\section{Related Works}
\subsection{SNN Training and Architecture Design}
The development of SNNs has long been hindered by the challenge of training non-differentiable binary spikes. To address this, researchers have focused on improving training methods and architectural designs. Recently, two primary methods for training high-performance SNNs have emerged. One approach is to convert ANNs into spike form through neuron equivalence \cite{li2021free,hao2023reducing,bu2022optimal}, known as ANN-to-SNN conversion. However, this method requires long simulation time steps and increases energy consumption. We employ the direct training method \cite{wu2018spatio} and apply surrogate gradient training. 
\par
Regarding architectural design, Spiking ResNet \cite{fang2021deep,shan2023or} has long dominated the SNN field because residual learning \cite{he2016deep} can address the performance degradation of SNNs as they deepen. The main differences among these are the locations of shortcuts and their ability to achieve identity mapping \cite{he2016identity}. Notably, MS-ResNet \cite{hu2024ms,qiu2024gated} maintains high performance while preserving the spike-driven sparse addition nature of SNNs by establishing residual connections between membrane potentials. Our E-3DSNN design draws on this idea and extends it to the 3D scene.
\vspace{-1mm}
\subsection{Feature Extractors on LiDAR-based 3D Recognition}
The key challenge in LiDAR-based 3D recognition is learning effective representations from sparse and irregular 3D geometric data. Currently, there exist two main approaches. Point-based methods \cite{qi2017pointnet,zhao2021point} utilize the PointNet series to directly extract geometric features from raw point clouds and make predictions. However, these methods require computationally intensive point sampling and neighbor search procedures. Additionally, in 3D scenes, numerous background points unrelated to the task contribute to redundant computations at each stage.
Voxel-based methods \cite{wu20153d,choy20194d,Voxelnet} first convert the point cloud into regular voxels and then use 3D sparse convolutions for feature extraction. Due to its efficiency advantages, this approach has been widely applied to various 3D tasks. Nevertheless, the improved accuracy is often accompanied by increased computational costs, limiting
its applicability in practical systems.
\par
Numerous studies \cite{lan2023efficient,ren2024spiking,wu2024point} in the SNN field combine spiking neurons with Point-based methods like PointNet. These methods have been successfully applied to simple datasets with shallow networks, but achieving high performance becomes more challenging as datasets and networks become larger and more complex, which restricts SNNs' application in 3D recognition. This is due to their oversight of SNNs' inherent sparsity, which can obscure local geometric information, and the high computational load of point-based methods, resulting in lengthy training times on large datasets.
We adopt a voxel-based approach for point cloud processing and leverage the sparse nature of spiking neurons to reduce unnecessary computation costs caused by 3D spatial redundancy.



\section{Efficient 3D Recognition SNN Backbone}
In this section, we begin by briefly introducing the spike neuron layer, followed by the Spike Voxel Coding (SVC)  and the Spike Sparse Convolution (SSC). Finally, we introduce our general efficient 3D recognition SNN backbone (E-3DSNN). 
\subsection{Leaky Integrate-and-Fire Spiking Neuron}
The Leaky Integrate-and-Fire (LIF) spiking neuron is the most popular neuron to balance bio-plausibility and computing complexity \cite{maass1997networks}. 
We first translate the LIF  spiking neuron to an iterative expression with
the Euler method (\citealt{wu2018spatio}), which can be described as:
\begin{align}
\label{eq:lif1}{U}^{t,n}&={H}^{t-1,n}+f({W^n},{X}^{t,n-1}), \\
\label{eq:lif2} {S}^{t,n}&=\Theta({U}^{t,n}-V_{th}),\\
    {H}^{t,n}& = \beta ({U}^{t,n}-{S}^{t,n}),
\label{eq:lif3}
\end{align}
where $\beta$ is the time constant, $t$ and $n$ respectively represent the indices of the time step and the $n$-th layer, $ W$ denotes synaptic weight matrix between two adjacent layers, $f(\cdot) $ is the function operation stands for convolution or fully connected layer, $ X$ is the input, and $\Theta(\cdot)$ is the Heaviside function. When the membrane potential $U$
exceeds the firing threshold $V_{th}$, the LIF neuron will trigger a spike $S$.
\par
However, converting the membrane potential of spiking neurons into binary spikes introduces inherent quantization errors, which significantly limit the model's expressiveness. To reduce the quantization error, we incorporate the Integer LIF (I-LIF) neuron \cite{luo2024integer} into our E-3DSNN, allowing us to rewrite Eq. \eqref{eq:lif2} as:
\begin{equation}
    {S}^{t,n}=\lfloor\operatorname{clip}\{{U}^{t,n},0,{D}\}\rceil,
    \label{ilif}
\end{equation}
where \(\lfloor \cdot \rceil\) denotes the rounding operator, \(\operatorname{clip}\{x, a, b\}\) confines \(x\) within range \([ a, b]\), and ${D}$ is a hyperparameter indicating the maximum emitted integer value by I-LIF. In the backpropagation stage, the  Eq. \eqref{eq:lif2} is non-differentiable. Previous studies have introduced various surrogate gradient functions \cite{wu2018spatio,neftci2019surrogate}, primarily designed to address binary spike outputs. In our approach, we consistently utilize rectangular windows as the surrogate function. For simplicity, we retain gradients only for neurons activated within the $[0, D]$ range, setting all others to zero. 
Moreover, I-LIF will emit integer values while training, and convert them into 0/1 spikes by expanding the virtual timestep to ensure that the inference is spike-driven with only sparse addition. 
\subsection{Spike Voxel Coding}
 In this section, we proposed a Spike Voxel Coding (SVC) scheme for efficiently transforming point clouds into spatio-temporal voxelized spike trains. The overall process of SVC processing a 3D point cloud is as follows.
 \par
 First, consider the input is a 3D point set with sparse voxelized 3D scene representation $\mathcal{V}=\{\mathcal{P},\mathcal{I}\}$. It contains voxels sets $V_{k}^{t}=\{P_{k}^{t},I_{k}^{t}\}$, where $P_{k}^{t}\in \mathbb{R}^{3}$ represents
the 3D coordinates and $I_{k}^{t} \in \mathbb{R}^{d}$ is the corresponding feature with $d$ channels at timestep $t$. Next, we divide the global voxel set $\mathcal{V}$ into $N$ non-overlapping voxel grids $[\mathcal{V}_{1}^{t},\mathcal{V}_{2}^{t},\ldots,\mathcal{V}_{N}^{t}],\mathcal{V}_{i}^{t}=\{V_{j}^{t}\mid {P}_{j}^{t}\in \Phi^{t}(i)\}$, where $\mathcal{V}_{i}^{t}$ represents the $i$-th voxel grid and $\Phi^{t}(i)$ means the index range of the $i$-th voxel grid at timestep $t$. Then we encode these voxel grids into spike trains, which can be expressed as:
\begin{equation}
    \mathcal{S}=\mathcal{SN}^{m}(\mathcal{F}^{m}(\mathcal{V})),
\end{equation}
where $\mathcal{SN}^{m}$ and $\mathcal{F}(\cdot)^{m}$  is $m$ consecutive I-LIF spiking neuron and sparse convolution, respectively. And $\mathcal{S}=[\mathcal{S}_{1}^{t},\mathcal{S}_{2}^{t},\ldots,\mathcal{S}_{N}^{t}]$ is a set of output spike trains. After our SVC, we obtain a sparse spiking representation $\mathcal{S}$ of the input 3D point cloud, which can reduce the storage requirements.

\subsection{Spike Sparse Convolution}
\paragraph{Vanilla Spike Convolution (VSC)} is performed on neuromorphic chips in a spike-driven manner. Then we will introduce how the VSC extracts 3D features. Consider weight $W^{t}$ contains $c_{\mathrm{in}} \times c_{\mathrm{out}}$ spatial kernels $K$ and $S^{t}_{p}$ as an input feature with $t$ timestep at position $p$ , VSC can be expressed by:
\begin{equation}
    U^{t}=\sum_{k\in K^{3}} W_{k}\cdot S^{t}_{\vec{p}_{k}},
    \label{vsc}
\end{equation}
Here $U^{t}$ is the output membrane potential and $\vec{p}_{k}$ is the position offset around center $p$, which can be expressed by:
\begin{equation}
    \vec{p}_{k}=p+k=(p_{x}+k_{x},p_{y}+k_{y},p_{z}+k_{z}),
\end{equation}
where  $k$ represents the kernel offset that enumerates all the discrete positions within the 3D kernel space  $K^{3}$.





\begin{figure}[t]
    \centering
    \includegraphics[width=1\linewidth]{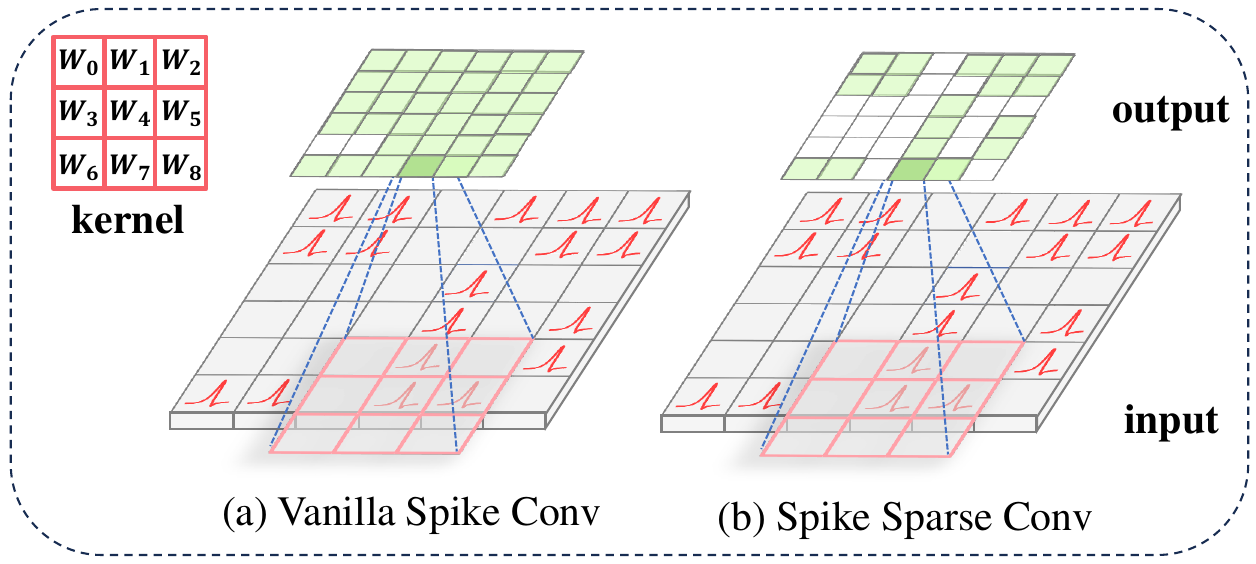}
    \caption{Comparison of Spike Sparse Conv (SSC) and Vanilla Spike Conv (VSC). Inputs and outputs are shown as 2D features for simplicity: green for activation, red for spikes, and white for no activation. On a neuromorphic chip, when a spike occurs, the address mapping function finds the synapses and neurons that need to be added and then takes out the corresponding weights to perform the addition operations. The only difference between VSC and SSC is the addressing mapping function. In SSC, it is specified that the convolution is performed only if there is a spike input at the position corresponding to $W_4$ (the center position of the convolution kernel). VSC does not have this restriction. }
    \label{fig:sparseconv}
\end{figure}

\paragraph{Spike Sparse Convolution (SSC)} VSC performs well in 2D scenes. However, in the 3D sparse point cloud, it needs to calculate each point and the computational complexity grows cubically when processing the point cloud, making it difficult to extract features efficiently. To address this issue, we propose Spike Sparse Convolution (SSC), which performs only on the key spiking locations, significantly reducing the computational requirements. It can be expressed by:
\begin{equation}
    U_{p}^{t}  = \sum_{k\in K^d}\alpha ({W}_{k} \cdot S^{t}_{\vec{p}_{k}}),
    \label{ssc}
\end{equation}
where  $\alpha \in \{0,1\}$ is a selector. When the center $ p \in S^{t} $ is the active binary spike, $\alpha$ equals 1, indicating that the position $p$ participates in the computation. $\alpha=0$ is the opposite. As depicted in Fig. \ref{fig:sparseconv}, SSC only has one more judgment condition than VSC when performing spike convolution. Given the commonality of VSC and SSC, we only need to slightly modify the addressing mapping function corresponding to VSC to execute SSC on the neuromorphic chip. 
 \par
 The specific process is as follows. Upon the reception of a spike, the SNN core first builds a rulebook, which records the activation spikes and the corresponding kernels space $K^3(p,P_{\mathrm{in}})$.
The kernel space is a subset of $K^{3}$, leaving out the empty
position. It is conditioned on the position $p$ and input feature
space $P_{\mathrm{in}}$ as:
\begin{equation}
    K^3(p,P_{\mathrm{in}})=\left\{k | p+k\in P_{\mathrm{in}},k\in K^3\right\}.
\end{equation}
Then the rulebook searches for and identifies the corresponding synaptic weights and the positions of the target neurons, and adds them together.
\paragraph{Computational Complexity} 
In this section, we analyze the computational complexity of our SSC by comparing its floating point operations (FLOPs) to those of VSC.
The FLOPs of VSC and SSC can be calculated by:
\begin{align}
\text{FLOPs}_{\text{VSC}}&= \sum_{l=1}^{L}2fr^{l}N_{l}k^{3}C_{l-1}C_{l},\\
\text{FLOPs}_{\text{SSC}}&=\sum_{l=1}^{L}2fr^{l} N_{l}^{r}C_{l-1}C_{l},
\end{align}
where $fr^{l}$ is the firing rate on layer $l$, $k$ is the kernel size of 3D convolution, $C_{l-1}$ is the input channels, $C_{l}$ is the output channels, $N_{l}$ is the output feature map size, and $N_{l}^{r}$ is the numbers of the rulebook, i.e. input and output activate points. As shown in Fig. \ref{fig:sparseconv}, SSC only has one more judgment condition than VSC when performing spike convolution. In SSC, the convolution is performed only if there is a spike input at the position corresponding to $W_4$ (the center position of the convolution kernel).

\subsection{Overall Architecture}
Fig. \ref{fig:main} shows the overview of our hieratical 3D Computer Vision SNN Backbone (E-3DSNN). Drawing inspiration of MS-ResNet \cite{hu2024ms}, we establish residual connections between the membrane potentials of spiking neurons. This avoids the common spike degradation issue \cite{yao2023spike} in SNNs and ensures that the network remains spike-driven during inference \cite{yaospike,yao2024scaling}. 
 Considering the input is a 3D point set with a sparse voxelized 3D scene representation $\mathcal{V}$, our E-3DSNN can be formulated as follows:
\begin{align}
   & \mathcal{S}^{t,0}= \mathcal{C}(\mathcal{V}),\\
    &U^{t,l}= \mathcal{B}^{l}(\text{Down}^{l}\{ \mathcal{S}^{t,0}\}),\\
    &U^{t,l+1}= \mathcal{B}^{l+1}(\text{Down}^{l+1}\{ U^{t,l}\}),
\end{align}
where $\mathcal{C}(\cdot)$ denotes spike voxel coding, and $l=1,\cdots,L$ represents the layer number, with $L$ equal to 4 in our study.
$\text{Down}(\cdot)$ is the downsample layer, which consists of a spiking neuron and a spike sparse convolution. Both the kernel size and stride are set to 2, reducing the spatial size to $\frac{1}{8}$ with each operation. $\mathcal{B}(\cdot)$ is the basic spike sparse block. Considering the input  of the basic spiking sparse block is $U$, this block  can be expressed as:
\begin{align}
   &U'=\text{SSC}\{\mathcal{SN}(U)\}+U,\\
   &U''=\text{SSC}^{m}\{\mathcal{SN}^{m}(U')\},
\end{align}

where $\text{SSC}^{m}$ and  $\mathcal{SN}^{m}$ indicate $m$ consecutive spike sparse convolution and spiking neurons, which is set to  2 in our study. The kernel size of SSC and stride are set to 3 and 1, respectively.
\subsection{Theoretical  Energy Consumption}
 The 3DSNN architecture can transform matrix multiplication into sparse addition, which can be implemented as an addressable addition on neuromorphic chips.
In the spike voxel coding layer, convolution operations serve as MAC operations that convert analog inputs into spikes, similar to direct coding-based SNNs \cite{wu2019direct}. Conversely, in SNN's architecture, the  Conv or FC layer transmits spikes and performs AC operations to accumulate weights for postsynaptic neurons. Additionally, the inference energy cost of E-3DSNN can be expressed as follows:
\begin{equation}
\label{energy}
    \begin{aligned}
E_{total}=E_{MAC}\cdot FL_{conv}^1+E_{AC}\cdot T \sum_{n=2}^N FL_{conv}^n \cdot fr^{n},
\end{aligned}
\end{equation}
where $N$ and $M$ are the total number of spike sparse conv, $E_{MAC}$ and $E_{AC}$ are the energy costs of MAC and AC operations, and $fr^{m}$, $fr^{n}$, $FL_{conv}^n$ and $FL_{fc}^m$ are the firing rate and FLOPs of the $n$-th spike sparse conv. And $E_{MAC}$ = 4.6pJ and $E_{AC}$ = 0.9pJ for various operations.

\begin{table*}[!t]
	\centering
 \begin{adjustbox}{max width=\linewidth} 
	\begin{tabular}{  c cc cc cc}
		\toprule
	Architecture	 &Method & Input 
   &$T \times D$
   & \begin{tabular}[c]{@{}c@{}}Param\\      (M)\end{tabular} 
   &\begin{tabular}[c]{@{}c@{}}Power\\      (mJ)\end{tabular}
   & \begin{tabular}[c]{@{}c@{}}Accuracy\\      (\%)\end{tabular} \\
		\midrule
		\multirow{5}*{ANN}&PointNet \cite{qi2017pointnet}\textit{\textsuperscript{CVPR}} & Point & N/A  & 3.47 &2.02& 89.2 \\
  &KPConv \cite{thomas2019kpconv}\textit{\textsuperscript{CVPR}} & Point & N/A  & 14.3&-& 92.9 \\
  & Pointformer \cite{zhao2021point}\textit{\textsuperscript{ICCV}}& Point  & N/A & 4.91 &5.1& 93.7 \\&3DShapeNets \cite{wu20153d}\textit{\textsuperscript{CVPR}}  & Voxel  & N/A & 6.92 &0.61& 77.3 \\
		&3DVGG-B \cite{graham2017submanifold} & Voxel & N/A  & 5.23 &0.45& 88.2 \\
  &E-3DSNN*& Voxel & N/A  &3.27 &0.30& 90.9 \\
\midrule
\multirow{5}*{SNN}
& SpikePointNet \cite{ren2024spiking}\textit{\textsuperscript{NeurIPS}} & Point  & 4$\times$1 & 3.47 &0.24& 88.2 \\

& SpikingPointNet \cite{lan2023efficient}\textit{\textsuperscript{ICCV}} & Point  & 16$\times$1& 3.47 &0.91& 88.6 \\
& P2SResLNet \cite{wu2024point}\textit{\textsuperscript{AAAI}}& Point  & 4$\times$1 & 14.3 &-& 88.7 \\
\cmidrule{2-7}
&\multirow{2}*{\textbf{E-3DSNN (Ours)}}& Voxel  &1$\times$4 & \textbf{1.87} &\textbf{0.02}& 91.5 \\
&& Voxel  &1$\times$4 & 3.27&0.04& \textbf{91.7} \\
		\bottomrule
	\end{tabular}
  \end{adjustbox}
  \caption{Shape classification results on the ModelNet40 dataset \cite{wu20153d}. Power is the estimation of energy consumption same as \cite{hu2024high,shan2024advancing}. * We convert 3.27M of E-3DSNN into
ANN with the same architecture. }
	\label{tab:modelnet40result}
\end{table*}

\section{Experiments}
In this section, we first give the hyper-parameters setting. Then we validate the E-3DSNN on diverse vision tasks, including 3D classification, object detection, and semantic segmentation.
Next, we ablate the different blocks of E-3DSNN to prove the effectiveness of our method. For further detailed information on architecture, more experiments on the NuScenes \cite{nuScenes} datasets, and visualizations, refer to the \textbf{Appendix}.
\subsection{Hyper-parameters Setting}
In this section, we give the specific hyperparameters of our training settings in all experiments, as depicted in Tab. \ref{tab:hyperparameters}. In this work, we train our E-3DSNN with 4 A100 GPUs.

\begin{table}[htbp]
    \centering
    \begin{adjustbox}{max width=\linewidth} 
    
    \begin{tabular}{lccc}
        \toprule 
         Parameter & ModelNet40 & KITTI &  SemanticKITTI  \\
         \midrule
         Learning Rate & $1e-1$  & $1e-2$ & $2e-3$  \\
         Weight Decay & $1e-4$  & $1e-2$ & $5e-3$  \\
         Batch Size & 16 & 64 & 96 \\
         Training Epochs & 200 & 80 & 100 \\
         Optimizer &SGD&Adam&AdamW\\
         \bottomrule
    \end{tabular}
      \end{adjustbox}
      \caption{Hyper-parameter training settings of 3DSNN.}
    \label{tab:hyperparameters}
\end{table}
\subsection{3D Classification} The ModelNet40 \cite{wu20153d} dataset contains 12,311 CAD models with 40 object categories. They are
split into 9,843 models for training and 2,468 for testing.  For the input data, we clip the point clouds to ranges of $[-0.2 \text{m}, 0.2 \text{m}]$ for the X-axis, $[-0.2 \text{m}, 0.2 \text{m}]$ for the Y-axis, and $[-0.2 \text{m}, 0.2 \text{m}]$ for the Z-axis. The input voxel size is set to $0.01 \text{m}$. In
terms of the evaluation metrics, we use the point cloud classification overall accuracy.
\par
As shown in Tab. \ref{tab:modelnet40result}, we compare
our method with the previous state-of-the-art ANN and SNN domain. Notably, with only 3.27M parameters, the E-3DSNN achieves the best accuracy of 91.7\%, regardless of voxel or point input in the SNN domain, showcasing significant advantages in both accuracy and efficiency. Specifically, E-3DSNN (This work) vs. SpikePointNet vs. SpikingPointNet: Power  0.02mJ vs. 0.24mJ vs. 0.91mJ; Param
1.87M vs. 3.47M vs. 3.47M; Accuracy 88.2\% vs. 88.6\% vs. 91.5\%.
That is, our model
has +2.8\% higher accuracy than SpikingPointNet \cite{lan2023efficient} with only
the previous 15.4\% energy consumption. Moreover, the performance gap between SNNs and ANNs is significantly narrowed. For instance, under lower parameters, the performance of Pointformer \cite{zhao2021point} and E-3DSNN  are comparable, and the energy efficiency is 127$\times$.

\begin{table*}[!t]
		\centering
			 \begin{adjustbox}{max width=\linewidth} 
				\begin{tabular}{ccccccccc}
					\toprule
					\multirow{2}{*}{Architecture}	 &\multirow{2}{*}{Method}&\multirow{2}{*}{Input}   &\multirow{2}{*}{$T \times D$} & \multirow{2}{*}{\begin{tabular}[c]{@{}c@{}}Param\\      (M)\end{tabular}}& \multirow{2}{*}{\begin{tabular}[c]{@{}c@{}}Power\\      (mJ)\end{tabular}} &
					\multicolumn{3}{c}{Car 3D AP (R11) }  \\
					&&&&& &Easy & Mod. & Hard \\
					\midrule
     \multirow{6}*{ANN}& PointRCNN \cite{PointRCNN}\textit{\textsuperscript{CVPR}}& Point&N/A& 4.0&22.5& 88.8  &78.6&77.3\\
     &PVRCNN \cite{PV-RCNN}\textit{\textsuperscript{CVPR}}& Point&N/A &13.1&34.9& 89.3   & 83.6&78.7  \\
     &Second \cite{SECOND}\textit{\textsuperscript{Sensor}}& Voxel &N/A&5.3&23.9& 88.6  &78.6&77.2 \\
     &VoxelRCNN \cite{Voxel-RCNN}\textit{\textsuperscript{AAAI}}& Voxel &N/A &7.5&28.9&89.4 &84.5&78.9\\
     &GLENet \cite{zhang2023glenet}\textit{\textsuperscript{IJCV}}& Voxel&N/A &8.3&-&89.8 &84.5& 78.7\\
     &$ \text{E-3DSNN}^{*}$  & Voxel&N/A &8.5&31.2&89.4 &83.7& 78.2\\
     \midrule
     \multirow{2}*{SNN}&$\text{SpikePointRCNN}^{\star}$& Point &1$\times$4& 4.0&4.4& 83.8  &72.1&71.9\\
     &\textbf{E-3DSNN (Ours)}& Voxel &1$\times$4 &8.5&\textbf{3.4}& \textbf{89.6}  &\textbf{84.0}&\textbf{78.7}\\
     				\midrule
				\end{tabular}
		\end{adjustbox}
		\caption{ 3D object detection results on the KITTI val benchmarks \cite{KITTI}. * We convert 8.5M of E-3DSNN into
ANN with the same architecture.}
		\label{tab:val_r11}
	\end{table*}
 
\begin{table*}[!t]
	\centering
 \begin{adjustbox}{max width=\linewidth} 
	\begin{tabular}{  c cc cc cc}
		\toprule
	Architecture	 &Method & Input 
   &$T \times D$
   & \begin{tabular}[c]{@{}c@{}}Param\\      (M)\end{tabular} 
   & \begin{tabular}[c]{@{}c@{}}Power\\      (mJ)\end{tabular} 
   & \begin{tabular}[c]{@{}c@{}}mIoU\\      (\%)\end{tabular} \\
		\midrule
		\multirow{5}*{ANN}&PointNet \cite{qi2017pointnet}\textit{\textsuperscript{CVPR}} & Point & N/A  & 3.5 &-& 14.6 \\
   &Pointformer V3 \cite{wu2024point}\textit{\textsuperscript{CVPR}} & Point  & N/A & 46.2&47.1 &72.3\\
  &SparseUNet \cite{graham2017submanifold}  & Voxel  & N/A & 39.1 &69.1& 63.8 \\
		&SphereFormer \cite{lai2023spherical}\textit{\textsuperscript{CVPR}}  & Voxel & N/A  & 32.3 &49.2& 67.8 \\
  
  &$ \text{E-3DSNN}^{*}$  & Voxel & N/A  & 20.1 &54.1& 69.4 \\
\midrule
\multirow{4}*{SNN}& $\text{SpikePointNet}^{\star}$& Point  & 1$\times$4 & 3.5 &-& 12.1 \\
& $ \text{SpikePointformer}^{\star}$ & Point  & 1$\times$4  &46.2&13.1&67.2 \\
\cmidrule{2-7}
&\multirow{2}*{\textbf{E-3DSNN (Ours)}}& Voxel  &1$\times$4 & \textbf{17.9} & \textbf{4.5}& 68.5\\
&& Voxel  &1$\times$4 & 20.1&6.1& \textbf{69.2} \\
		\bottomrule
	\end{tabular}
  \end{adjustbox}
  \caption{3D semantic segmentation results on Semantic KITTI val benchmarks \cite{behley2019semantickitti}. * We convert 20.1M of E-3DSNN into
ANN with the same architecture.}
	\label{tab:kittisegresult}
\end{table*}

\subsection{3D Object Detection} The  large KITTI dataset \cite{geiger2012we}
consists of 7481 training samples, which are divided into trainsets with 3717 samples and validation sets with 3769 samples. In detection, E-3DSNN are evaluated as backbones equipped with VoxelRCNN Head \cite{Voxel-RCNN}. We transform  OpenPCDet \cite{openpcdet2020} codebase into a spiking version and use it to execute our model. Raw point clouds are divided into regular voxels before being input to our 3DSNN on KITTI \cite{KITTI}. For the input data, we clip the point clouds to the following ranges: $[0, 70.4] \text{m}$ for the X-axis, $[-40, 40] \text{m}$ for the Y-axis, and $[-3, 1] \text{m}$ for the Z-axis. The input voxel size is set to $(0.05 \text{m}, 0.05 \text{m}, 0.1 \text{m})$. In terms of the evaluation metrics, we use the Average Precision (AP) calculated by 11 recall positions for the Car class.
\par
As shown in Tab. \ref{tab:val_r11}, we compare our method in 3D object detection with the previous state-of-the-art (SOTA) ANN domain. Since no SNN has yet reported results on the KITTI dataset, we employ the I-LIF spiking neuron \cite{luo2024integer} to directly convert the PointRCNN \cite{PointRCNN} architecture into a spike-based version, referred to as SpikePointRCNN. We obtained 89.6\%, 84.0\%, 78.7\% AP, which is higher than the prior state-of-the-art SNN  by a large margin, i.e., 5.8\%, 11.9\%, 6.8\% absolute improvements on easy, moderate, and hard levels of class Car.
E-3DSNN also has significant advantages over existing SNNs and ANNs regarding parameters and power. For instance,  E-3DSNN (This work) vs. SpikePointNetRCNN vs. VoxelRCNN: AP 89.6\% vs. 83.8\% vs. 89.4\%; Power  3.4mJ vs. 4.4mJ vs. 28.9mJ. In summary, E-3DSNN achieved state-of-the-art performance in the SNN domain in terms of both accuracy and efficiency on the KITTI dataset, while also achieving results comparable to ANNs.

\subsection{3D Semantic Segmentation}  The large SemanticKITTI dataset \cite{behley2019semantickitti}  consists of sequences from the raw KITTI dataset, which contains 22 sequences in total. Each sequence includes around 1,000 lidar scans, corresponding to approximately 20,000 individual frames. We first transform Pointcept \cite{pointcept2023} codebase into a spiking version and use it to execute our model. 
Then we design an asymmetric encoder-decoder structure similar to UNet \cite{choy20194d,wu2023masked}, with our E-3DSNN as encoder responsible for extracting multi-scale features and the decoder sequentially fusing the extracted multi-scale features with the help of skip connections. For voxelize implementation, the window size is set to $[120 \text{m}, 2^\circ, 2^\circ]$ for $(r, \theta, \phi)$. During data preprocessing, we restrict the input scene to the range $[-51.2 \text{m}, -51.2 \text{m}, -4 \text{m}]$ to $[51.2 \text{m}, 51.2 \text{m}, 2.4 \text{m}]$. The voxel size is set to $0.1 \text{m}$. 
\par
As shown in Tab. \ref{tab:kittisegresult}, we compare
our method in 3D Semantic Segmentation with the previous state-of-the-art ANN domain. Since no SNN has yet reported results on the SemanticKITTI dataset, we employ the I-LIF spiking neuron \cite{luo2024integer} to convert the PointNet and Pointformer architectures into spike-based versions directly.
We found that our E-3DSNN achieves the best mIoU of 69.2\%, which is 2.0\% and 1.6\% higher than the previous SOTA  SNN.
Our E-3DSNN also demonstrates significant advantages over existing SNNs and ANNs in terms of parameter efficiency and power consumption for 3D semantic segmentation.
For instance, E-3DSNN (This work) vs. SpikePoinformer vs. SphereFormer: mIoU 69.2\% vs. 67.2\% vs. 67.8\%; Power  8.2mJ vs. 19.0mJ vs. 49.2mJ; Param: 20.1M vs. 46.2M vs. 32.3M.

\begin{figure*}[!t]
    \centering
    \includegraphics[width=0.90\linewidth]{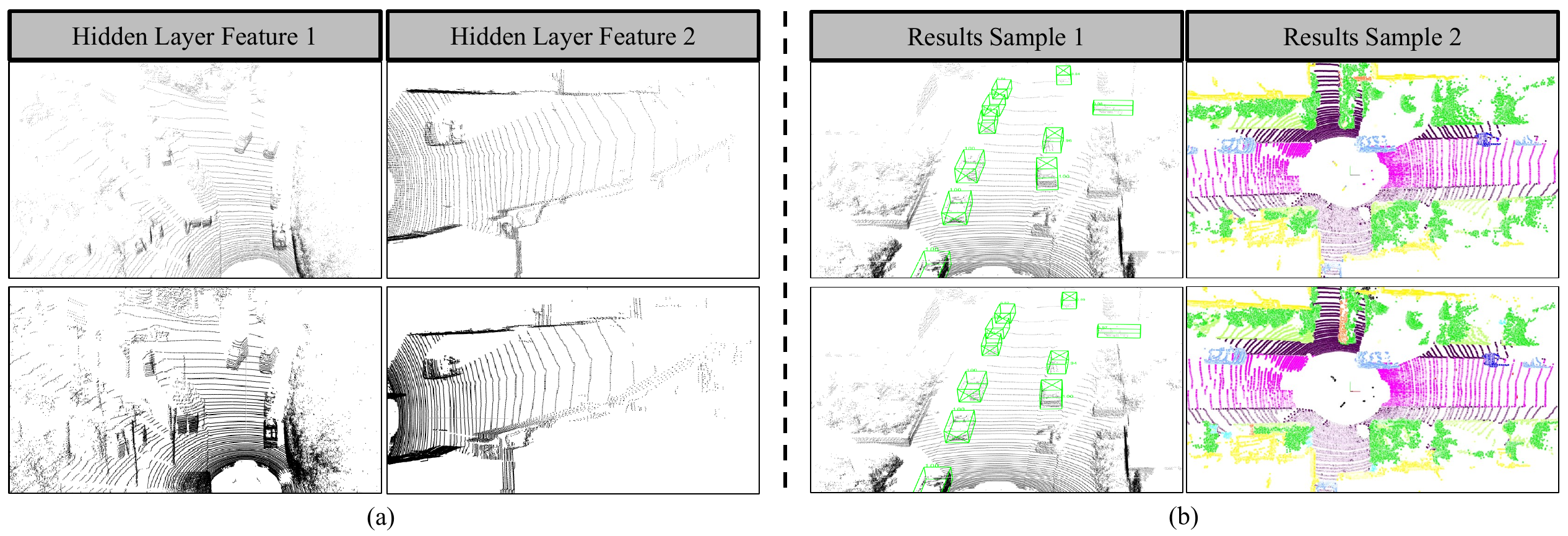}
    \caption{Visualization of E-3DSNN in hidden layer features and results. (a) We compared the hidden layer features generated with (top) and without SVC and SSC (bottom).
(b) We compared the results of our E-3DSNN (top) in detection and segmentation with the ground truth (bottom).}
    \label{fig:vis}
\end{figure*}
\subsection{Ablation Study}
We first compared the results of ANN and SNN with the same architecture on the Semantic KITTI validation benchmarks. As shown in Tab. \ref{tab:ablation}, while E-3DSNN’s mIoU accuracy is 0.2\% lower than the corresponding ANN, it shows an 8.8$\times$ improvement in energy efficiency. This indicates that SNNs have significant potential in efficiently processing sparse 3D point clouds.
\par
Next, we ablate two components of our E-3DSNN, namely the SVC and SSC, to verify the effectiveness of the proposed method. As shown in Tab. \ref{tab:ablation}, using SVC alone yields a slight decrease in mIoU by 0.3\% but achieves a 1.8× improvement in energy efficiency. When both SVC and SSC are employed, there is a 0.2\% reduction in mIoU, but energy efficiency improves by 8.8×.  Therefore, the proposed SSC and SVC can significantly reduce power consumption and improve efficiency while maintaining high performance. Their combined use yields even more substantial improvements, highlighting their effectiveness in enhancing energy efficiency in 3D recognition tasks.

\par
Then we evaluate the effects of varying $T$ and $D$. We found that increasing the number of timesteps can enhance performance but affect inference time and increase power consumption. For instance, for the set with $D$ = 1, extending $T$ from 2 to 4 increases the mIoU by 1.4\% but doubles the power consumption. Additionally, we found that expanding $D$ while keeping $T$ fixed improves performance and reduces power consumption. For instance, 2 × 1 vs. 2 × 2: mIoU, 67.1\% vs. 68.9\%; Power, 8.4mJ vs. 8.1mJ. 1 × 2 vs. 1 × 4: mIoU, 67.9\% vs. 69.2\%; Power, 6.3mJ vs. 6.1mJ.

\begin{table}[htbp]
\centering
\begin{adjustbox}{max width=\linewidth} 
\begin{tabular}{cccccc}
\toprule
Method    &   SVC &SSC  &$T\times D$  & \begin{tabular}[c]{@{}c@{}}Power\\      (mJ)\end{tabular}  & \begin{tabular}[c]{@{}c@{}}mIoU\\      (\%)\end{tabular}\\ \midrule
 $ \text{ANN}^{*}$ &-& -   & N/A     &54.1& 69.4  \\
  \cmidrule{1-6}
 \multirow{6}*{E-3DSNN}&\cmark &-    &$1\times 4$&29.1& 69.3     \\
 &\cmark &\cmark     &$1\times 4$ &6.1 &   69.2\\ 
 \cmidrule{2-6}
  &\cmark &\cmark     &$1\times 2$ &6.3 &   67.9\\ 
 &\cmark &\cmark    &$2\times 1$   & 8.4&67.1\\
 &\cmark &\cmark    &$2\times 2$   & 8.1&68.9\\
 &\cmark &\cmark    &$4\times 1$   &16.1 &68.5\\
    \bottomrule
\end{tabular}
\end{adjustbox}
\caption{Ablation study of the E-3DSNN  on Semantic KITTI val benchmarks \cite{behley2019semantickitti}.  * We convert 20.1M of E-3DSNN into
ANN with the same architecture.}
\label{tab:ablation}
\end{table}

\subsection{Visualization}
We evaluated the effectiveness of E-3DSNN in reducing irrelevant redundant points in the background. By training E-3DSNN with and without SVC and SSC on Semantic KITTI \cite{behley2019semantickitti} and KITTI \cite{KITTI} datasets, we generated the hidden layer feature maps and final detection and segmentation results shown in Fig. \ref{fig:vis}. It can be observed that our SSC and SVC help 3D SNNs significantly reduce irrelevant redundant points in the background in 3D detection and segmentation tasks. For instance, in the intermediate feature maps Fig. \ref{fig:vis} (a) and (b), we notice that most foreground points are preserved, while road points, being easily identifiable as redundant, are largely removed. In the result Fig. \ref{fig:vis} (c) and (d), we observe that our E-3DSNN achieves visual effects in detection and segmentation that are comparable to those of ANNs with the same architecture. For instance, in detection, our E-3DSNN has detected all car categories with high confidence. In segmentation, for fine-grained categories such as fence and sidewalk, our E-3DSNN demonstrates excellent segmentation performance.

\section{Conclusion}
This paper significantly narrows the performance gap between ANN and SNN on 3D recognition tasks. We accomplish this through two key issues with SNN in processing 3D point cloud data. First, to tackle the disordered and uneven nature of point cloud data, we propose the Spike Voxel Coding (SVC) scheme, which significantly improves storage and preprocessing efficiency. Second, to overcome the rapid increase in computational complexity when applying SNNs to 3D point clouds, we introduce Spike Sparse Convolution (SSC), which reduces redundant computations on background points.  The E-3DSNN backbone utilizes these innovations along with residual connections between membrane potentials to handle various 3D computer vision tasks efficiently. Experiments conducted on ModelNet, KITTI, and Semantic KITTI datasets demonstrate that E-3DSNN achieves state-of-the-art performance in terms of accuracy and efficiency across different tasks including 3D classification, object detection, and semantic segmentation. We hope our investigations pave the way for efficient 3D recognition and inspire the design of sparse event-driven SNNs.

\bibliography{aaai25}
\newpage

\section{Appendix}
In the appendix, we first provide details of the architecture used in our backbone. Next, we present the experimental results on the large-scale NuScenes dataset \cite{nuScenes}, followed by our detection and segmentation performance on the KITTI \cite{KITTI} and Semantic KITTI \cite{behley2019semantickitti} datasets.

\section{More Experiments on Large Dataset NuScenes}
The NuScenes dataset \cite{nuScenes} consists of 1,000 driving sequences, divided into 700 for training, 150 for validation, and 150 for testing. Collected using a 32-beam LIDAR and 6 cameras, the dataset includes annotations for 10 classes. In our ablation study, detection models are trained on a quarter of the training data and evaluated on the full validation set. We evaluate our E-3DSNN as a backbone with the VoxelNext Head \cite{chen2023voxelnext}, transforming the OpenPCDet \cite{openpcdet2020} codebase into a spiking version for execution. The raw point clouds are voxelized before being input to our 3DSNN on the NuScenes dataset \cite{nuScenes}, with point clouds clipped to [-54m, 54m] for the X and Y axes, and [-5m, 3m] for the Z axis. The default voxel size is (0.075m, 0.075m, 0.2m).
\par
As shown in Tab. \ref{tab:kittisegresult}, our E-3DSNN demonstrates significant advantages over existing SNNs and ANNs in terms of parameter efficiency and power consumption for 3D object detection.
Despite a slight performance loss on the large-scale NuScenes dataset \cite{nuScenes} compared to the state-of-the-art VoxelNext, our E-3DSNN significantly reduces energy consumption to one-tenth of the original requirement. For instance, E-3DSNN (This work) vs. VoxelNext vs. CenterPoint: NDS 64.1 vs. 66.6 vs. 64.5; Power 3.1mJ vs. 33.1mJ vs. 35.1mJ; Parameters: 6.9M vs. 7.9M vs. 8.1M. We conducted experiments comparing the performance of SNNs and ANNs with the same architecture. We observed that E-3DSNN's NDS is only 1.3\% lower than the corresponding ANN while demonstrating a 9.5x improvement in energy efficiency. This indicates that SNNs offer significant potential for low-power processing of point clouds.

\begin{table}[htbp]
    \centering
\begin{adjustbox}{max width=\linewidth} 
    \begin{tabular}{cccccc} 
    \toprule
\multirow{2}{*}{Type}  &\multirow{2}{*}{Blocks} &\multirow{2}{*}{ Channels}&Param & mIoU \\
    &&&(M)&(\%)\\
    \midrule
    E-3DSNN (T)&$[1,1,1,1]$&$[16, 32, 64, 128]$&1.8&62.5\\
    E-3DSNN (S)&$[1,1,1,1]$&$[24,48,96,160]$&3.2&64.3\\
    E-3DSNN (B)&$[2,2,2,2]$&$[16,32,64,128]$&9.8&68.5\\
    E-3DSNN (L)&$[2,2,2,2]$&$[64, 128, 128,256]$&12.1&69.2\\
    \bottomrule 
    \end{tabular}
    \end{adjustbox}
    \caption{Comparison between various versions of E-3DSNN's encoder. The timestep and output bit is set to $1\times4$.}
    \label{tab:type_comparison}
\end{table}
\begin{table}[!h]
	\centering
 \begin{adjustbox}{max width=\linewidth} 
	\begin{tabular}{   cc c cccc}
		\toprule
	Method & Input 
   &$T \times D$
   & \begin{tabular}[c]{@{}c@{}}Param\\      (M)\end{tabular} 
   & \begin{tabular}[c]{@{}c@{}}Power\\      (mJ)\end{tabular} 
    &NDS
   \\
		\midrule
		PointPillar \cite{PointPillars} & Point & N/A  & 6.5 & 31.9&58.2 \\
  CenterPoint \cite{CenterPoint} & Point  & N/A & 8.1 &35.1&64.5\\
 VoxelNext \cite{chen2023voxelnext}  & Voxel & N/A  & 7.9 & 33.1&66.6\\
  $ \text{E-3DSNN}^{*}$  & Voxel &1$\times$4   & 6.9 & 29.5&65.4 \\
  \midrule
\textbf{E-3DSNN (Ours)}& Voxel  &1$\times$4 & \textbf{6.9} & \textbf{3.1}&\textbf{64.1}\\

		\bottomrule
	\end{tabular}
  \end{adjustbox}
  \caption{3D object detection results on the nuScenes val benchmarks \cite{nuScenes}.  * We convert 6.9M of E-3DSNN into
ANN with the same architecture.}
	\label{tab:kittisegresult}
\end{table}
\section{Architecture details}
We present multiple versions of E-3DSNN, achieved by adjusting the number of blocks and channels in each stage. The impact on performance and efficiency is demonstrated in Tab. \ref{tab:type_comparison}, where all models are evaluated on four A100 with Semantic KITTI dataset \cite{behley2019semantickitti} to ensure a fair comparison. We applied the encoders from E-3DSNN (T) and E-3DSNN (S) to classification on ModelNet40 and detection on KITTI, respectively, while the more complex E-3DSNN (B) and E-3DSNN (L) were used for segmentation on Semantic KITTI. It was observed that increasing the number of channels in the model, while keeping the number of blocks constant, improved the model's performance by 1.8\%. Additionally, increasing the number of blocks while keeping the channels constant resulted in a 6.0\% performance boost. This indicates that the number of layers has the most significant impact on E-3DSNN's performance.

\section{Visualization}
\begin{figure*}[htbp]
    \centering
    \includegraphics[width=0.95\linewidth]{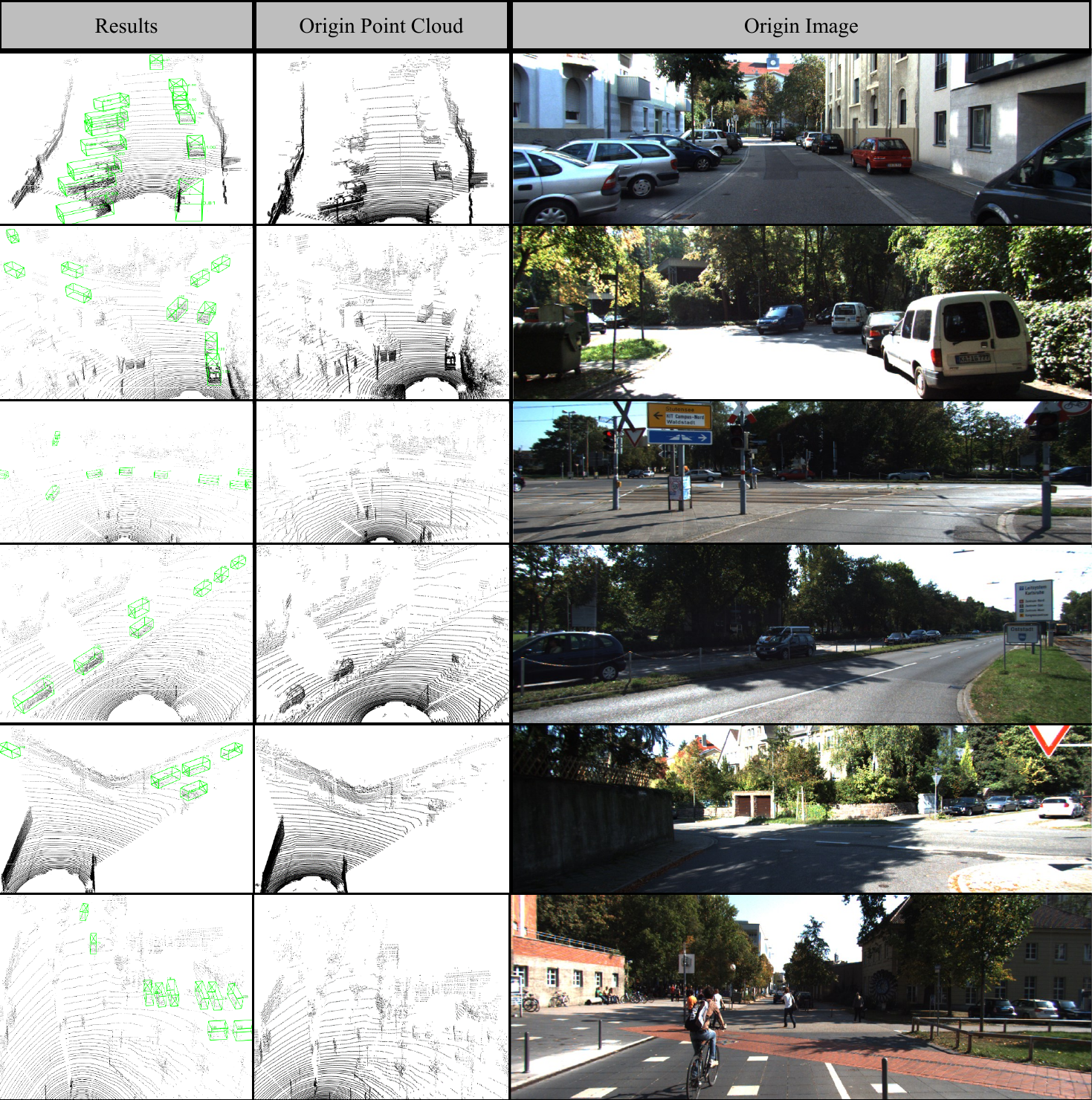}
    \caption{ Visualization of results on KITTI dataset \cite{KITTI}. Our E-3DSNN excels in the 3D object detection task}
    \label{fig:KITTI}
\end{figure*}

\begin{figure*}[htbp]
    \centering
    \includegraphics[width=0.95\linewidth]{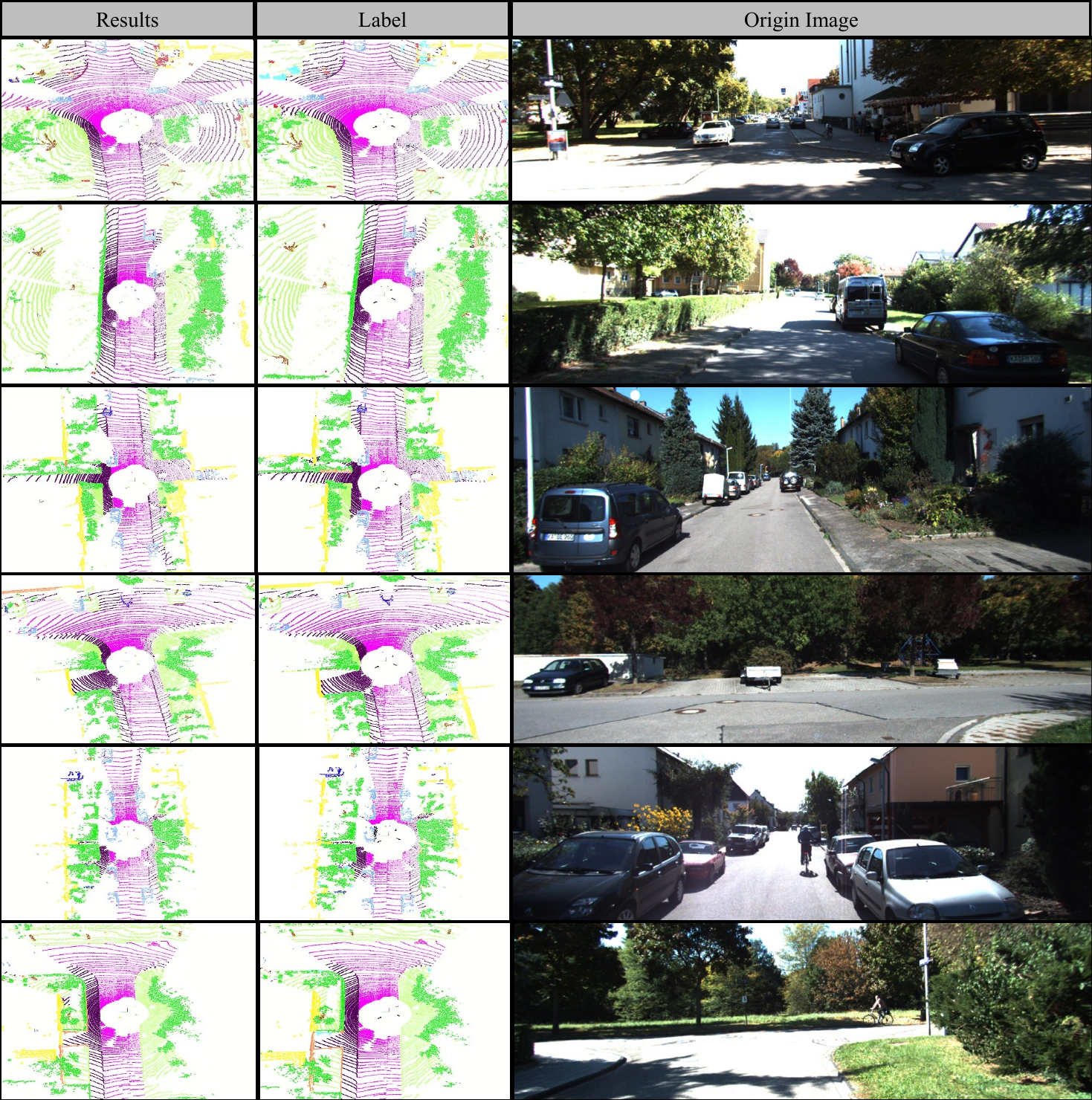}
    \caption{Visualization of results on Semantic KITTI dataset \cite{behley2019semantickitti}. Our E-3DSNN excels in the 3D semantic segmentation task.}
    \label{fig:Semantic}
\end{figure*}

As shown in Fig. \ref{fig:KITTI} and \ref{fig:Semantic}, we present the detection and segmentation results of E-3DSNN on the KITTI and Semantic KITTI datasets to demonstrate the effectiveness of our approach.

\end{document}